\documentclass[11pt,a4paper]{article}
\usepackage[hyperref]{emnlp2018}
\usepackage{times}
\usepackage{latexsym}
\usepackage{amsmath}
\usepackage{graphicx}
\usepackage{multirow}
\usepackage{url}

\aclfinalcopy 

\title{A Recipe for Arabic-English Neural Machine Translation}

\author{Abdullah Alrajeh \\
  National Center for AI and Big Data Technology \\
  King Abdulaziz City for Science and Technology (KACST) \\
  Riyadh, Saudi Arabia \\
  {\tt asrajeh@kacst.edu.sa}}

\date{}

\begin{document}
\maketitle
\begin{abstract}
In this paper, we present a recipe for building a good Arabic-English
neural machine translation. We compare neural systems with traditional 
phrase-based systems using various parallel corpora including UN, ISI 
and Ummah. We also investigate the importance of special preprocessing 
of the Arabic script. The presented results are based on test sets from 
NIST MT 2005 and 2012. The best neural system produces a gain of +13 BLEU 
points compared to an equivalent simple phrase-based system in NIST MT12
test set. Unexpectedly, we find that tuning a model trained on the whole 
data using a small high quality corpus like Ummah gives a substantial 
improvement (+3 BLEU points). We also find that training a neural system with
a small Arabic-English corpus is competitive to a traditional phrase-based
system.

\end{abstract}

\section{Introduction}
Neural networks succeed to show impressive results as part of a statistical
machine translation (SMT) system in the work of \citet{devlin-EtAl}. Since then, 
the research shifted more towards an end-to-end approach. Currently, neural machine 
translation (NMT) has become the dominant approach in the field achieving 
state-of-the-art results in many translation tasks \cite{bojar-EtAl:2017:WMT1}.

\citet{Junczys-Dowmunt16c} investigate 30 translation directions using the UN corpus 
(around 335M words). The experiments, based on test sets from the same corpus, show 
that NMT is superior to the traditional approach (i.e. phrase-based SMT). One of the 
investigated task is translation between Arabic and English but without special 
preprocessing for Arabic. Large improvement is observed (around 3 BLEU points) over 
phrase-based SMT in both directions.

\citet{AlmahairiCHC16} compared a neural system against phrase-based one in Arabic-English
translation task and found them to be comparable based on NIST 2005 test set. It is also observed 
that NMT is superior to SMT in an out-of-domain test set. In all cases, preprocessing
of Arabic script did improve the translation quality.

In this paper, we further investigate Arabic to English translation using several corpora
including Ummah, ISI, UN and many others. We compare the performance of NMT against 
phrase-based SMT. In our experiments, we applied Arabic preprocessing, which includes 
normalization and tokenization, to see its impact on both NMT and SMT systems. Our results 
is based on NIST MT sets for the year 2005, 2006 and 2012.

In the next section, we give a brief introduction to neural machine translation. Section 3
lists the parallel corpora will be used and show some statistics. Section 4 presents our SMT 
and NMT experiments followed by the conclusion.

\section{Neural Machine Translation}
\citet{kalchbrenner-blunsom:2013:EMNLP} set the foundation of neural machine translation by
proposing an end-to-end encoder-decoder approach. Convolutional neural networks (CNN) are used
to encode a source sentence then generates its translation by recurrent neural networks (RNN).

Long sentences propose a challenge for RNN where there are long distance reordering.
\citet{NIPS2014_5346} develop sequence-to-sequence models that use RNN for both encoding 
and decoding. Standard RNN units are replaced with Long Short-Term Memory (LSTM) units to 
capture long-term dependencies. \citet{cho-EtAl:2014:EMNLP2014} introduce Gated Recurrent 
Unit (GRU) which is simpler than LSTM. 

In the previous work, a source sentence is encoded into a fixed-length vector which is
a bottleneck holding NMT from being competitive to SMT particularly in long sentences.
\citet{BahdanauCB14} introduce the powerful attention mechanism that allows the decoder
to focus on different words while translating.

These advancements and other such as byte pair encoding \cite{sennrich-haddow-birch:2016:P16-12},
to achieve open vocabulary NMT, pave the way for new state-of-the-art translation systems.

Mathematically, the probability of a translation sentence (${\bf y}:y_1,\dots,y_I$) of 
an input sentence (${\bf x}:x_1,\dots,x_J$) is computed as follows:
\begin{align}
\centering
&p({\bf y}|{\bf x}) = \prod^I_{i=1} p(y_i|y_1,\dots,y_{i-1},{\bf x}) \\
&p(y_i|y_1,\dots,y_{i-1},{\bf x}) = g(y_{i-1},s_i,c_i)
\end{align}

where $s_i=f(s_{i-1},y_{i-1},c_i)$ is a hidden state in the RNN decoder while $c_i$ is the context
vector computed from all hidden states ($h$) in the RNN encoder as follows: 
\begin{align}
\centering
&c_i=\sum^J_{j=1} \alpha_{ij} h_j \\
&\alpha_{ij} = \frac{\exp(a(s_{i-1},h_j))}{\sum^J_{k=1}\exp(a(s_{i-1},h_k))}
\end{align}

where $h_j=f(h_{j-1},x_j)$, $\alpha_{ij}$ is its weight and $a(.)$ is an alignment model shows 
the importance of the input word $j$ in translating the output word $i$. This mechanism allows the 
decoder to just pay attention to the related input words. Note that, the function $f$ that produces 
the next hidden state in the encoder and decoder can be defined as LSTM or GRU.

Usually, an input sentence is encoded by a forward RNN but a backward RNN, that reads the sentence in a 
reverse order, is found to improve the performance \cite{NIPS2014_5346}. A bidirectional RNN also has
been successful \cite{BahdanauCB14}. It reads the sentence in both directions then concatenates the 
forward and backword hidden states as follows:
\begin{equation}
\centering
h_j=[\overrightarrow{h}^T_j;\overleftarrow{h}^T_j]^T
\end{equation}

\section{Corpora}
There are many parallel corpora available for building Arabic-English translation systems. The UN 
corpus\footnote{https://conferences.unite.un.org/uncorpus} is an obvious choice for many researchers 
and will be used in our experiments. It is composed of parliamentary documents of the United Nations 
since 1990 for Arabic, Chinese, English, French, Russian, and Spanish \cite{ZIEMSKI16.1195}.

There are also 11 LDC\footnote{http://ldc.upenn.edu/} corpora have been selected. These include Ummah and ISI
with catalogue numbers LDC2004T18 and LDC2007T08, respectively. Ummah corpus contains news stories while ISI
was extracted automatically from Arabic Gigaword and English Gigaword. The rest are mostly from GALE project
with catalogue numbers LDC2004T17, LDC2005T05, LDC2008T09, LDC2009T09, LDC2013T10, LDC2013T14, LDC2015T05, 
LDC2015T07 and LDC2015T19.

Besides that, we used all Arabic-English corpora available on OPUS\footnote{http://opus.nlpl.eu/} website
\cite{TIEDEMANN12.463,Rafalovitch,WOLK2014126}. We exclude MultiUN because we already have larger version. 
OpenSubtitles and Tanzil are also excluded due to their low quality.

Table~\ref{ar-en-corpora} shows statistics of all corpora. The total number of all English words is close to 
half a billion words.

\begin{table}[h]
\centering
\resizebox{\columnwidth}{!}{%
\begin{tabular}{|c|c|c|cc|} \hline
No. & Corpus & Sentences & \multicolumn{2}{c|}{Ar-En words} \\ \hline
1   & Ummah		 & 80k	 & 2.3m  & 2.9m  \\
2   & ISI		 & 1.1m  & 28.9m & 30.8m \\
3   & \small{LDC2004T17} & 19k & 441k & 581k \\
4   & \small{LDC2005T05} & 5k & 106k & 135k \\
5   & \small{LDC2008T09} & 3k & 55k & 68k \\       
6   & \small{LDC2009T09} & 10k & 145k & 198k \\       
7   & \small{LDC2013T10} & 8k & 182k & 240k \\
8   & \small{LDC2013T14} & 5k & 89k & 124k \\              
9   & \small{LDC2015T05} & 18k & 285k & 379k \\       
10  & \small{LDC2015T07} & 20k & 330k & 440k \\       
11  & \small{LDC2015T19} & 6K & 156k & 210k \\
12  & OPUS 		 & 639k & 13.8m & 13.8m \\       
13  & UN 		 & 185m & 398m & 448m \\ \hline
\end{tabular}
}
\caption{Statistics of all Arabic-English corpora (m: million, k: thousand).}
\label{ar-en-corpora}
\end{table}

\section{Experiments}
We present SMT and NMT results on Arabic-English based on NIST MT sets for 
the year 2005, 2006 and 2012 (see Table~\ref{nist-sets}). As commonly used 
in machine translation, we evaluated the translation performance by BLEU 
score \citep{Papineni}.

\begin{table}[h]
\centering
\resizebox{\columnwidth}{!}{%
\begin{tabular}{|c|cc|cc|} \hline
          & \multicolumn{2}{c|}{Ar-En Sentences} & \multicolumn{2}{c|}{Ar-En words} \\ \hline
MT06 (dev)  & 1797   & 1797 & 42k & 54k  \\
MT05 (test) & 1056   & 4224 & 26k & 130k \\
MT12 (test) & 1378   & 5512 & 35k & 191k \\ \hline
\end{tabular}
}
\caption{Statistics of NIST MT sets.}
\label{nist-sets}
\end{table}

The systems are 
trained on different datasets ranging from small to very large. Training corpora
in Table~\ref{ar-en-corpora} are grouped into 4 sets:
\begin{itemize}
\item Set A: Ummah corpus
\item Set B: Ummah, ISI and LDC2004T17
\item Set C: all corpora except UN
\item Set D: all corpora
\end{itemize}

The reasons for this setting are the following.
Low-resource MT task is a known challenge for NMT \cite{koehn-knowles:2017:NMT}.
We would like to see if this is the case for Arabic-English task (Set A). 
\citet{AlmahairiCHC16} report the first result on Arabic NMT therefore Set B are 
chosen for comparison. Finally, UN corpus might add no benefit \cite{devlin-EtAl}
since it is not the news domain (Set C and D).

{\bf Preprocessing} \, In our experiments, we applied Arabic preprocessing, which includes
normalization and tokenization (ATB scheme), to see its impact on both SMT and NMT systems.
We used Farasa \cite{DBLP:conf/naacl/AbdelaliDDM16}, a fast Arabic segmenter.
The maximum sentence length is 100.

{\bf Phrase-based MT} \, We use Moses toolkit \citep{Koehn2007} with its default settings. 
The language model is a 5-gram built from the English side with interpolation and Kneser-Ney 
smoothing \citep{Kneser} built by KenLM \cite{heafield-EtAl:2013:Short}. Word alignments are 
extracted by {\tt fast\_align} \cite{dyer-chahuneau-smith:2013:NAACL-HLT}. We tune the system using 
MERT technique \cite{Och2003b}. The chosen option for the reordering model is {\tt msd-bidirectional-fe}.

{\bf Neural MT} \, We use Marian, an efficien and fast NMT system written in C++ \cite{junczys2018marian}.
The system has implemented several models. The {\tt s2s} option is chosen which is equivalent to Nematus 
models \cite{sennrich-EtAl:2017:EACLDemo} that are RNN encoder-decoder based with attention mechanism.
The basic training script provided is used. To achieve open vocabulary, we apply byte pair encoding (BPE)
\cite{sennrich-haddow-birch:2016:P16-12} setting the maximum size of the joint Arabic-English vocabulary to 90,000.

Table~\ref{smt} reports BLEU scores of many phrase-based SMT systems trained on various datasets.
Clearly, preprocessing of the Arabic side is important. Substantial gain is observed when the training
data is small (Set A). Note that adding UN corpus to the training data improves the BLEU score as in set D.

\begin{table}[h]
\centering
\resizebox{\columnwidth}{!}{%
\begin{tabular}{|c|l|c|c|c|} \hline
Set & System & MT05 & MT12 & avg\\ \hline
\multirow{2}{*}{A} & baseline 		& 39.49 & 22.50 & 31.00\\
		   & + ar preprocessing  & 42.17 & 31.87 & 37.02\\ \hline
\multirow{2}{*}{B} & baseline  		& 49.71 & 34.25 & 41.98\\ 
                   & + ar preprocessing  & 51.65 & 37.06 & 44.35\\ \hline
\multirow{2}{*}{C} & baseline           & 51.32 & 38.12 & 44.72\\
                   & + ar preprocessing  & 52.76 & 40.80 & 46.78\\ \hline
\multirow{2}{*}{D} & baseline           & 52.57 & 40.02 & 46.30\\
                   & + ar preprocessing  & 53.45 & 41.11 & 47.28\\ \hline
\end{tabular}
}
\caption{BLEU scores of Arabic-English SMT.}
\label{smt}
\end{table}

Table~\ref{nmt} presents NMT systems's performance in BLEU. Compared to Table~\ref{smt}, NMT is
superior to SMT in all cases. The best NMT system produces a gain of +13 BLEU points in NIST MT12 
test set. Unexpectedly, NMT is similar or better than SMT even with a small dataset
(Set A is less than 3 million English words). Note that the gap in BLEU between NMT and SMT increases
with more training data. Arabic preprocessing improves the performance as in Table~\ref{smt} which
indicates that BPE is not sufficient. We find that tuning a model trained on the whole data using
a high quality corpus like Ummah (Set A) gives us a substantial improvement. Finally, 
an independent ensemble of 5 best models boosts the score with +1.5 BLEU.

\begin{table}[h]
\centering
\resizebox{\columnwidth}{!}{%
\begin{tabular}{|c|l|c|c|c|} \hline
Set & System & MT05 & MT12 & avg\\ \hline
\multirow{2}{*}{A} & baseline           & 41.62 & 19.47 & 30.55\\
                   & + ar preprocessing & 44.15 & 31.86 & 38.01\\ \hline
\multirow{2}{*}{B} & baseline           & 53.27 & 38.19 & 45.73\\
                   & + ar preprocessing & 54.69 & 40.07 & 47.38\\ \hline
\multirow{2}{*}{C} & baseline           & 57.02 & 45.73 & 51.38\\
                   & + ar preprocessing & 58.35 & 47.04 & 52.70\\ \hline
\multirow{2}{*}{D} & baseline           & 57.31 & 45.81 & 51.56\\
                   & + ar preprocessing & 58.43 & 47.74 & 53.09\\
                   & + tuning		& 61.26 & 52.53 & 56.90\\ 
                   & + ensemble of 5 	& 62.98 & 54.27 & 58.63\\ \hline
\end{tabular}
}
\caption{BLEU scores of Arabic-English NMT.}
\label{nmt}
\end{table}

Training the whole data on a single GPU took 4 days\footnote{NVidia GTX 1080 Ti, CPU 4.20GHz and Hard disk SSD}.
The disk size of the best model is just 645 MB. It is very compact compared to the phrase table alone in 
SMT which is 8.5 GB.

During the experiments, other models have been tried like {\tt transformer} \cite{NIPS2017_7181}
but no improvement is gained. It is also the case for the joint vocabulary's size.

\section{Conclusion}
We present Arabic to English machine translation using various training datasets.
We compare neural systems with traditional ones (i.e. phrase-based SMT). We also
investigate the importance of special preprocessing of the Arabic script. The systems
are tested on NIST MT 2005 and 2012.

After the experiments, we draw the following conclusions. In both NMT and SMT systems, Arabic preprocessing 
improves the translation quality as found by \citet{AlmahairiCHC16}. Although UN corpus is not in the news 
domain, a gain is observed in both systems. Neural MT is superior to phrase-based MT in all cases.
NMT able to perform very well given a small corpus. Finally, tuning a model trained on the whole data using
a small high quality corpus (i.e. Ummah) gives a substantial improvement. The best NMT system produces a gain 
of +13 BLEU points in NIST MT12 test set.

There are techniques we have not considered in this work but might improve the translation quality such as
back translation \cite{sennrich-haddow-birch:2016:P16-11}.

\bibliography{emnlp2018}
\bibliographystyle{acl_natbib_nourl}

\end{document}